\pdfoutput=1
\documentclass[11pt]{article}

\usepackage{acl}

\usepackage{enumitem}
 
\usepackage{multicol}
\usepackage{booktabs} 
\usepackage{multirow}
\usepackage{graphicx}
\usepackage{times}
\usepackage{makecell}
\usepackage{latexsym}
\usepackage[T1]{fontenc}
\usepackage[utf8]{inputenc}
\usepackage{microtype}
\usepackage{inconsolata}
\usepackage{amsmath}
\usepackage{amssymb}
\usepackage{makecell}
\usepackage{array}
\usepackage{tikz}
\usepackage[font=small,labelfont=bf]{caption}

\usepackage{bm}
\usepackage{tcolorbox}

\definecolor{myGreen}{RGB}{127,210,85}
\definecolor{myOrange}{RGB}{242,154,66}
\definecolor{myYellow}{RGB}{247,223,65}
\definecolor{myRed}{RGB}{232,80,43}
\definecolor{myViolet}{RGB}{162,57,102}
\definecolor{myBlue}{HTML}{4686f3}
\definecolor{myYellowv2}{HTML}{E6C802}
\definecolor{myOrangev2}{HTML}{ED8E55}
\definecolor{MyGreenv2}{HTML}{009B55}
\definecolor{MyRedv2}{HTML}{c22f2f}
  
\newcommand{\positiveevaluationmetric}{\textsc{Accuracy-w/-Gold}}  
\newcommand{\negativeevaluationmetric}{\textsc{Accuracy-w/o-Gold}}  
\newcommand{\newevaluationmetric}{\textsc{OmniAccuracy}} 
\newcommand{\newetask}{\textsc{Classify-w/o-Gold}}  
\newcommand{\tasknameequation}{\textsc{EquInfer}}  
\newcommand{\tasknamebanking}{\textsc{Bank-77}}  
\newcommand{\tasknamemctest}{\textsc{MC-Test}}  
\newcommand{\taskname}{\textsc{Know-No}}  
\newcommand{\hintoption}{\textsc{Hint-as-Option}}  
\newcommand{\hintinstru}{\textsc{Hint-in-Instru}}  
\newcommand{\hintno}{\textsc{No-Hint}}  
\newcommand{\noneofthem}{\texttt{none-of-them}}  

\title{LLMs' Classification Performance is Overclaimed}

\date{}

\author{
Hanzi Xu$\textsuperscript{*}$ \ \ \ 
Renze lou$\textsuperscript{$\S$}$ \ \ \ 
Jiangshu Du$\textsuperscript{$\dagger$}$ \ \ \ 
Vahid Mahzoon$\textsuperscript{*}$ \ \ \ 
Elmira Talebianaraki$\textsuperscript{*}$ \\
\textbf{Zhuoan Zhou$\textsuperscript{*}$ \ \ \ 
Elizabeth Garrison$\textsuperscript{*}$ \ \ \
Slobodan Vucetic$\textsuperscript{*}$ \ \ \ 
Wenpeng Yin$\textsuperscript{$\S$}$}\\
$\textsuperscript{*}$Temple University \ \ \ $\textsuperscript{$\dagger$}$University of Illinois at Chicago \ \ \ $\textsuperscript{$\S$}$Penn State University\\
 \texttt{\{hanzi.xu, slobodan.vucetic\}@temple.edu} \ \ \ 
\texttt{wenpeng@psu.edu}
}

\begin{document}
\maketitle
\begin{abstract}
In many classification tasks designed for AI or human to solve, gold labels are typically included within the label space by default, often posed as ``which of the following is correct?'' This standard setup has traditionally highlighted the strong performance of advanced AI, particularly top-performing Large Language Models (LLMs), in routine classification tasks. However, when the gold label is intentionally excluded from the label space, it becomes evident that LLMs still attempt to select from the available label candidates, even when none are correct. This raises a pivotal question: Do LLMs truly demonstrate their intelligence in understanding the essence of classification tasks?

In this study, we evaluate both closed-source and open-source LLMs across representative classification tasks, arguing that the perceived performance of LLMs is overstated due to their inability to exhibit the expected comprehension of the task. This paper makes a threefold contribution: i) To our knowledge, this is the first work to identify the limitations of LLMs in classification tasks when gold labels are absent. We define this task as \newetask~ and propose it as a new testbed for LLMs. ii) We introduce a benchmark, \taskname, comprising two existing classification tasks and one new task, to evaluate \newetask. iii) This work defines and advocates for a new evaluation metric, \newevaluationmetric, which assesses LLMs' performance in classification tasks both when gold labels are present and absent\footnote{Our code, data, and raw results will be publicly available at \url{https://github.com/xhz0809/Know-No}}.

\end{abstract}                     
\section{Introduction}

\begin{figure}[t]
	\begin{center}
		\centering
		\includegraphics[width=0.47\textwidth]{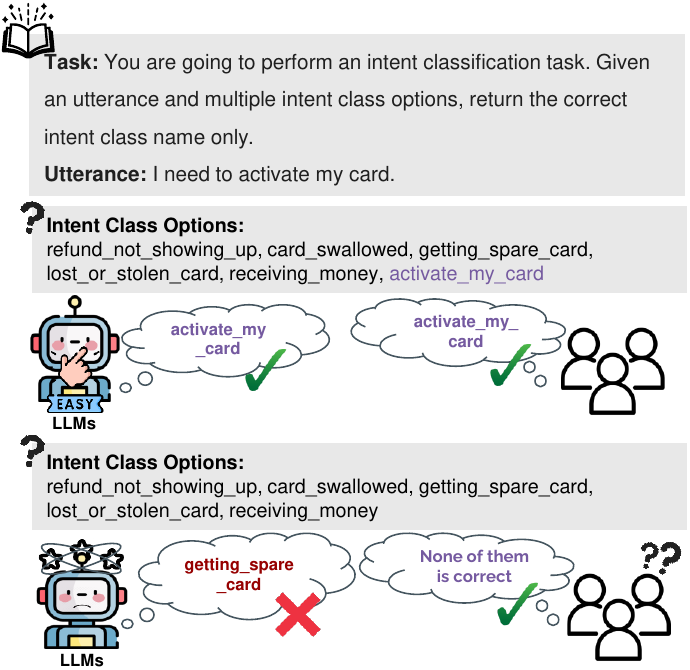}
        \end{center}
  \vspace{0em}
\caption{Latest LLMs (GPT-4o, claude-3-opus, and gemini-1.5-pro as of June 29, 2024\protect\footnotemark) vs. Human when the gold label is present or absent. }
\label{fig:GPT_Know-No_example}
\vspace{-1em}
\end{figure}

Let's begin with the example in Figure~\ref{fig:GPT_Know-No_example}, which illustrates the use of the latest LLMs for a straightforward classification problem.

\footnotetext{API access with temperature set as 0}

In this simple scenario, GPT-4o, the latest large language model (LLM), performs comparably to humans when the correct answer is included among the options. However, interestingly, even this advanced LLM does not show uncertainty by indicating ``no correct answer'' or ``all options seem incorrect'', a behavior consistently demonstrated by humans when the correct answer is not provided.

Why should we be concerned about this particular phenomenon, especially in the era of LLMs? There are two primary reasons: i) Given the versatility of LLMs, they can process inputs with any set of labels by following natural language instructions, even when the correctness of the labels is unknown. The expected behavior from LLMs should mirror that of humans in the previous example: identifying correct labels when present or indicating the absence of correct ones without risking users accepting false responses. In contrast, traditional classifiers, trained on fixed label sets, are limited to predicting within those specific labels and lack the flexibility to handle open sets of labels. ii) LLMs are predominantly designed as generative models, prioritizing the enhancement of their generative capabilities, often at the expense of discriminative capabilities \cite{DBLP:conf/emnlp/SunL0WGZ023}. Many researchers argue that classification tasks are perceived as easy for LLMs, as evidenced by their consistently high performance \cite{DBLP:conf/emnlp/SunL0WGZ023, DBLP:journals/corr/abs-2306-09719,DBLP:journals/corr/abs-2402-07470}. However, the above example raises the question of whether the performance of LLMs in classification tasks has been overstated due to current evaluation benchmarks and metrics only capturing incomplete human behavior.

To investigate this question, we present three standard classification tasks as benchmarks: \tasknamebanking~ (intent classification task, \citealp{DBLP:journals/corr/abs-2003-04807}) , \tasknamemctest~ (multiple-choice question answering task, \citealp{DBLP:conf/emnlp/RichardsonBR13}) , and \tasknameequation, a newly assembled task where the objective is to infer the correct equation from four candidates given surrounding paragraphs in scientific papers. This benchmark, termed \taskname, encompasses classification tasks with inputs of varying lengths, label sizes, and label scopes, including instance-level and task-level label spaces.

We define a novel evaluation metric, \newevaluationmetric, designed to accurately assess the human-level discrimination intelligence of LLMs in classification tasks. This metric integrates the performance of LLMs across two dimensions within the \taskname~framework: i) \positiveevaluationmetric: representing the conventional accuracy when the correct label is provided. ii) \negativeevaluationmetric: indicating the accuracy when the correct label is not provided. We argue that \newevaluationmetric~offers a more comprehensive reflection of LLMs' classification performance.

In summary, our contributions can be outlined as follows:
\begin{itemize}
    \item To the best of our knowledge, this is the first study to uncover the limitations of LLMs in classification tasks when gold labels are absent. We designate this task as \newetask~and propose it as a novel evaluation framework for LLMs.
    
    \item We introduce a benchmark, \taskname, which encompasses two established classification tasks alongside a newly devised one, aimed at evaluating \newetask.
    
    \item This study introduces and advocates for a new evaluation metric, \newevaluationmetric, tailored to assess LLMs in classification tasks. This metric combines performance metrics when gold labels are both present and absent, offering a more comprehensive evaluation of LLMs' capabilities.

\end{itemize}


\begin{table*}[!ht]
\centering
\setlength{\tabcolsep}{2pt}
 \begin{tabular}{l|lrcccc}
  & \#input & $|\mathrm{input}|$ & label format   & \#label   &  label scope & challenge \\
 \hline
 
 \tasknamebanking~\citep{DBLP:journals/corr/abs-2003-04807} & 1,000 &  12  & phrase           & 77   & task-level  & moderate \\
 \tasknamemctest~\citep{DBLP:conf/emnlp/RichardsonBR13}    & 1,000 &  220  & phrase\&sent & 4    &  instance-level & low \\
 \tasknameequation   & 1,049 &  1,925  & latex equation  & 4    & instance-level & high
 \end{tabular}
 \vspace{-0.5em}
 \caption{Statistics of \taskname. ``Task-level label scope'' refers to a dataset where all instances share the same set of labels. In contrast, ``instance-level label scope'' entails instances having varying label sizes and options, as illustrated in the example in the Introduction. The ``challenge'' categorization depends on whether domain expertise or common knowledge suffices for the task.}
 \label{tab:data}
 \vspace{-1em}
\end{table*} 

\section{Related Work} 

\paragraph{LLMs' high performance on classification tasks.}
It has been a trend to use LLM generation to solve classification problems, either as standalone classification tasks \cite{DBLP:conf/emnlp/SunL0WGZ023, DBLP:journals/corr/abs-2306-09719,DBLP:journals/corr/abs-2402-07470} or mixed with other NLP tasks in multi-task learning \cite{DBLP:conf/icml/LongpreHVWCTZLZ23,DBLP:conf/emnlp/WangMAKMNADASPK22, DBLP:conf/acl/MishraKBH22}. Remarkable performance metrics from GPT have been observed, including an average accuracy above 90\% on five well-known NLP benchmark text classification datasets (SST-2, AGNews, R8, R52, MR) reported in \cite{DBLP:conf/emnlp/SunL0WGZ023} with zero-shot prompting. Additionally, \citet{DBLP:journals/corr/abs-2306-09719} demonstrate 95-98\% accuracy in sentiment analysis (SST-2/IMDB/Yelp), over 93\% in semantic role labeling (CoNLL2009), and 92-98\% in part-of-speech identification (Penn, WSJTweets) with few-shot demonstrations. As the latest LLMs are seen as reliable solutions for NLP classification, their true understanding of the essence of the classification task has not been properly evaluated.

\paragraph{LLMs' Challenges in Task Comprehension}

It is essential to investigate the rationale behind the model’s predictions and determine the extent to which we can trust its output \cite{DBLP:journals/scirobotics/GunningSCMSY19, DBLP:journals/natmi/Rudin19}. Recent studies have raised concerns about whether LLMs truly understand the tasks they perform despite their good performance. 

Many studies have shown that LLMs can achieve promising performance when they are asked to provide step-by-step reasoning in their answers \cite{DBLP:conf/nips/Wei0SBIXCLZ22, DBLP:conf/nips/ZelikmanWMG22, DBLP:journals/corr/abs-2210-06726}. However, LLM-generated reasoning has been found to be unfaithful despite its apparent effectiveness \cite{DBLP:conf/nips/TurpinMPB23,DBLP:journals/corr/abs-2307-13702}. The performance boost might be attributed to the extra computation provided by the explanation tokens \cite{DBLP:conf/nips/Wei0SBIXCLZ22,DBLP:journals/corr/abs-2307-13702}. An inspection of the reasoning rationales generated by the model reveals that they often fail to make logical sense \cite{DBLP:conf/nips/ZelikmanWMG22}. It is common to see rationales that simply repeat content from the question without providing a reasonable explanation. Many of the rationales fail to effectively support the claims or address the reasoning required, indicating that the model often does not truly understand the content and reasoning behind the question, even if it arrives at the correct answer.

Similarly, recent works have been evaluating the cognition-inspired intelligence of LLMs by testing latest LLMs on NLP generation tasks to evaluate their capabilities across multiple dimensions, including reading comprehension, commonsense reasoning, discourse comprehension, and paragraph/document-level understanding \cite{DBLP:journals/corr/abs-2403-00126, DBLP:journals/corr/abs-2301-06627}. When the problem-solving process is decomposed into three sub-steps: knowledge recall, knowledge utilization, and answer generation, the results reveal that, despite high scores in answer generation performance, the score for knowledge utilization is significantly lower, by up to 34\%. Additionally, they point out that LLMs' proficiency in language processing does not necessarily translate to a similar level of cognitive capability if looking at the correlation between different capabilities, revealing the current shortcomings of LLMs in true understanding.

\section{Approach}

\subsection{\taskname~Benchmark} \label{sec:datasets}

In this work, we collect representative classification datasets to cover (i) multiple task types and difficulty levels, and (ii) various label sizes and label scopes (task-level label space or instance-level label space). 
Specifically, we build this benchmark ``\taskname'' with two existing datasets \tasknamebanking~ \citep{DBLP:journals/corr/abs-2003-04807}, \tasknamemctest~ \citep{DBLP:conf/emnlp/RichardsonBR13}, along with a new dataset proposed by us, named \tasknameequation. An overview of the \taskname~statistics is provided in Table \ref{tab:data}. We will first briefly introduce \tasknamebanking~ and \tasknamemctest, followed by a detailed explanation of the construction process for \tasknameequation.

\paragraph{\tasknamebanking~\cite{DBLP:journals/corr/abs-2003-04807}.} Inputs are simple sentences (customer service queries) in the banking and financial domain, all sharing the same label space of 77 intents (``task-level label''). The original dataset comprises 13,083 inputs; however, due to budget limitations, we randomly selected 1,000 instances for this study, ensuring coverage of all 77 labels. We chose this dataset because of its moderate difficulty level and large label size.

\paragraph{\tasknamemctest~\cite{DBLP:conf/emnlp/RichardsonBR13}.} \tasknamemctest~is a pioneering multiple-choice reading comprehension benchmark. It includes 660 elementary-school-level stories, each accompanied by four multiple-choice questions with four unique answer options for each question (``instance-level'' label). We randomly selected 250 stories (1000 questions) as our test set. We chose \tasknamemctest~ because of its simplicity (most questions can be answered by keyword matching rather than deep reasoning), which may help us uncover surprising behaviors of the latest LLMs.

\begin{figure*}[h]
    \centering
    \begin{tcolorbox}[colframe=black, colback=white, width=\textwidth, boxrule=0.5mm]
        \textbf{Context before:} \\
        ``[$\cdots\cdots\cdots$] Therefore, the positive sample is the corresponding augmented sentence, while the negative samples are the augmented versions of other original source sentences from the same mini-batch. $\bm{e}_{x^i}$ and $\bm{e}_{z^i}$ are the average representations along the sequence dimension from the encoder outputs. Apart from the contrastive loss, the standard cross-entropy loss is calculated as:''
        \\
        \textbf{Context after:} \\
        ``We combine both losses as the final loss:
        \begin{align}
        \mathcal{L} = \mathcal{L}_{ce} + \lambda \mathcal{L}_{ctr} \nonumber
        \end{align}
        where $\lambda$ is an interpolation factor. We incorporate the augmented source inputs $z$ to ensure that the model can still generate correct translations with noisy input. [$\cdots\cdots\cdots$]''
        \\
        \textbf{Equation options:} \\
        \begin{tabular}{ll}
            A: $\mathcal{L}_{ce} = -\sum_{i=1}^{N} \sum_{c=1}^{C} y_{i,c} \log(p_{i,c})$ &
            B: $\mathcal{L}_{ce} = - \sum _{i=1}^{N} (\log P_{\theta}(y^i | x^i) + \log P_{\theta}(y^i | z^i))$ \\
            \vspace{0mm} \\
            C: $\mathcal{L}_{ce}  = - \frac{1}{N} \sum_{i=1}^{N} \sum_{c=1}^{C} y^i_c \log(p^i_c)$ &
            D: $\mathcal{L}_{ce} = -\sum_{i=1}^N \sum_{k=1}^C y_k^i \log \hat{y}_k^i$ 
        \end{tabular}
    \end{tcolorbox}
      \vspace{-1em}
    \caption{An example of \tasknameequation, where the equation labeled with ``B'' is correct.}
    \label{fig:example_dataset}
\end{figure*}

\begin{table*}[!ht]
    \centering
    \begin{tabular}{c m{1cm}|l}
      \multirow{3}{*}{\parbox{0.5cm}{\centering w/ \\$\mathcal{G}$}} &  &  \begin{minipage}[t]{0.82\linewidth}
    \texttt{Instruction}: For the following input and options, please return the correct option(s). \\ \texttt{Input}: $\cdots$ \\ \texttt{Options}: $ID_1$. [$option_1$] ; $ID_2$. [$option_2$]; $\cdots$; \textcolor{red}{$ID_{\mathcal{G}}$. [$option_{\mathcal{G}}$]}; $\cdots$
  \end{minipage} \\\hline
  
      \multirow{10}{*}{\parbox{0.5cm}{\centering w/o \\$\mathcal{G}$}}  & \multirow{3}{*}{\parbox{1cm}{hint as \\option.}} &  \begin{minipage}[t]{0.82\linewidth}
    \texttt{Instruction}: For the following input and options, please return the correct option(s). \\ \texttt{Input}: $\cdots$ \\ \texttt{Options}: $ID_1$. [$option_1$] ; $ID_2$. [$option_2$]; $\cdots$ \textcolor{blue}{$ID_n$. none-of-them}
  \end{minipage} \\\cline{2-3}

       & \multirow{4}{*}{\parbox{1cm}{hint in instru.}}  &  \begin{minipage}[t]{0.82\linewidth}
    \texttt{Instruction}: For the following input and options, please return the correct option(s), or return ``\textcolor{blue}{none-of-them}'' if you believe none of the options is correct. \\ \texttt{Input}: $\cdots$ \\ \texttt{Options}: $ID_1$. [$option_1$] ; $ID_2$. [$option_2$]; $\cdots$ 
  \end{minipage} \\\cline{2-3}

       & \multirow{3}{*}{\parbox{1cm}{no hint}} &  \begin{minipage}[t]{0.82\linewidth}
    \texttt{Instruction}: For the following input and options, please return the correct option(s). \\ \texttt{Input}: $\cdots$ \\ \texttt{Options}: $ID_1$. [$option_1$] ; $ID_2$. [$option_2$]; $\cdots$
  \end{minipage} \\
    \end{tabular}
    \caption{Prompting LLMs in \taskname. $\mathcal{G}$: gold label (i.e. ``$ID_{\mathcal{G}}$. [$option_{\mathcal{G}}$]'', in red). Blue: \noneofthem~ hint. In addition to the content above, we append ``Your answer:'' as the suffix of the prompt to ensure that the model correctly understands the task. Sample prompt for each dataset can be found in the Appendix~\ref{appendix:prompt_example}}
    \label{tab:prompts}
      \vspace{-1em}
\end{table*}

\paragraph{\tasknameequation.} We design this task to mimic the paper reviewing process, where reviewers must determine if an equation is valid based on its context. This task requires intensive domain expertise.

i) \textbf{Data Crawling}. We crawled a total of 4,951 papers' LaTeX source packages from ArXiv, focusing on papers accepted by top-tier NLP conferences.\footnote{ACL, EMNLP, NAACL, TACL, and EACL, etc.} We excluded papers that were unsuitable for this task, including 1) papers without any LaTeX equations and 2) papers with overly complicated equations (e.g., equations with nested structures or custom commands). This filtering process resulted in 1,449 papers. From each paper, we randomly sampled up to 3 equations, leading to a final set of 3,877 equations.

ii) \textbf{Task Formulation}. We formulate this task as a multiple-choice classification, where each instance includes 1000-word context before the equation and 1000 words after, all in the original LaTeX format (details on scaling the optimal context length can be found in Appendix~\ref{appendix:equation_scale}). The model must select the correct LaTeX equation from one positive option (the gold equation from the original paper) and three negative options.

iii) \textbf{Label Space Construction}. To craft high-quality negative options, we mask out the target gold equation in a paper and prompt \textsc{GPT-4} to generate the masked equation based on the context before and after the equation (100 words on each side).\footnote{We also provide GPT-4 with the left part of the ``\(=\)'' sign from the gold equation to make the LLM-crafted negative equations more similar to the gold equation, thereby increasing the challenge.}

iv) \textbf{Quality Control (Automatic and Manual)}. We filtered out negative equations if: a) they were identical to the gold equation; b) GPT-4 could easily recognize the flaws. For the latter, we provided GPT-4-Turbo with the negative equation and asked whether the equation had significant flaws. The remaining negative equations are thus ``hard'' options that can easily deceive the LLMs.\footnote{In practice, for each classification instance, if any of the three negative equations can fool GPT-4, we will keep all three negative equations in our dataset. This ensures there is at least one ``hard'' negative equation in each classification instance.} Among the original 3,877 instances, the instance will be abandoned if it cannot gather 3 qualified negative equations. The above process results in a total of 1,449 instances. Since all the above filtering steps are based on LLMs that may still leave some false negative equations, we asked humans to further filter out classification instances with any suspicious false negative equations (i.e., LLM-crafted negative equations that are logically correct). After human filtering, we have a total of 1,049 classification instances.

An example instance of \tasknameequation~is shown in Figure \ref{fig:example_dataset}.

\subsection{Prompting LLMs}\label{sec:prompt}

Here we elaborate on our prompts for scenarios where gold labels are present and where gold labels are absent, respectively.

\paragraph{Prompt when the gold label is present (w/ $\mathcal{G}$).} The prompt used can be seen in the first block of Table \ref{tab:prompts}.

\paragraph{Prompt when the gold label is absent (w/o $\mathcal{G}$).} In this case, the gold option is deleted. The key question is whether we should provide hints for the LLMs on how to handle situations where all options appear incorrect, and how to implement these hints. In this work, based on real-world scenarios, we design three types of hints (blocks 2-4 in Table \ref{tab:prompts}):
\begin{itemize}
    \item \hintoption: Even though no gold label is available, we provide ``none-of-them'' as one of the options. This mirrors common human behavior when no valid options are found, leading to the selection of this choice. An answer will only be counted as correct when LLMs choose ``\noneofthem''.
    \item \hintinstru: In contrast to the above hint type, here we do not include ``\noneofthem'' as an option. Instead, in the instruction, we explicitly request the LLM to output ``\noneofthem'' if no correct option is found. An answer will only be counted as correct when LLMs return ``\noneofthem''.
    \item \hintno: No hint at all. The instruction is the same as ``w/ $\mathcal{G}$'' except for the absence of the gold label.

    The evaluation of \hintno~is more complicated: based on our observations, some LLMs, especially top-performing ones, tend to generate a new option with an explanation if they believe no correct options are provided (this may also indicate data leakage of our datasets in LLM pretraining, which we will analyze in $\mathcal{Q}_4$ of Section \ref{sec:analysis}). This type of LLM response creates two challenges: i) it lacks a fixed format, making automatic parsing for system evaluation infeasible; ii) in reality, if an LLM response contains a reasonable label for the input, it requires us to understand the task and conduct some reasoning, which is beyond the scope of automatic processing. Also, we want to avoid using LLM for evaluation, which might introduce even more errors. Therefore, for \hintno, we always report human performance by manually reviewing LLM responses.
\end{itemize}

We believe that any of the hint types mentioned above would work for human users. Using diverse prompts, we aim to comprehensively evaluate the model's performance without the gold option and avoid behavior specific to a particular prompt.

\subsection{\newevaluationmetric: A new evaluation metric}

Our goal with \newevaluationmetric~is to have it reflect model performance both when the gold label is present and absent. Therefore, we define \newevaluationmetric~in the following straightforward form:
\begin{equation}
    \newevaluationmetric = \frac{1}{2}\cdot(\mathcal{A}_{w/}+ E[\mathcal{A}_{w/o}])
\end{equation}
where $\mathcal{A}_{w/}$ represents the accuracy when the gold label is present, and $E[\mathcal{A}_{w/o}]$ indicates the expectation of accuracy when the gold label is absent. In practice, $E[\mathcal{A}_{w/o}]$ can be achieved by combining multiple prompting techniques, such as the three hints styles described in Section \ref{sec:prompt}. We also encourage researchers to explore the most appropriate form for their particular research. In this work, we adopt the following form:
\begin{equation}
    E[\mathcal{A}_{w/o}] = \frac{\sum_{i=1}^n\mathcal{A}_{w/o}^i}{n}
\end{equation}
i.e., we take the average performance across all conditions where no correct answer is presented, including ``\hintoption'', ``\hintinstru''~ and ``\hintno'', as the comprehensive and robust assessment of LLMs when the gold label is missing.

\begin{table*}[h]
    \centering
    \begin{tabular}{cc p{2.2cm}|c|c|c|c|c|c}
    \hline\hline
    & & & \multicolumn{2}{c|}{Closed-source} & \multicolumn{3}{c|}{Open-source} & \multirow{2}{*}{Human} \\
   & &  & GPT-4 & Claude3 & Llama3 & Gemma & Mistral & \\\hline
   
        \multirow{6}{*}{\tasknamemctest}  & \multicolumn{2}{c|}{w/ $\mathcal{G}$ (i.e., $\mathcal{A}_{w/}$)} &98.67  & 98.26&94.23 & 39.0 &87.53 & 100.00 \\\cline{2-9}
         & \multirow{4}{*}{w/o $\mathcal{G}$} & Hint as option &80.17 &49.83 &43.10 &4.33 &39.97 & 96.00\\\cline{3-9}
         & & Hint in instru.  &80.40 &62.17 &3.83 &15.26 &30.27 & 97.00\\\cline{3-9}
         & & No hint  & 41.30 & 60.30 & 50.10 & 15.90 & 33.60 & 93.00 \\\cline{3-9}
         & & \enspace\enspace\enspace $E[\mathcal{A}_{w/o}]$  &67.29 &57.43  &32.34  &11.83  &34.61 &95.33\\\cline{2-9}
         & \multicolumn{2}{c|}{\newevaluationmetric} &\textbf{82.98} &\textbf{77.85}  &\textbf{63.29}  &\textbf{25.41}   &\textbf{61.07}  & \textbf{97.67}\\\hline

       \multirow{6}{*}{\tasknamebanking}  & \multicolumn{2}{c|}{w/ $\mathcal{G}$ (i.e., $\mathcal{A}_{w/}$) } & 69.40  &65.75  &42.53  &39.03  &45.1 & --\\\cline{2-9}
         & \multirow{4}{*}{w/o $\mathcal{G}$} & Hint as option & 1.83 &0.9 &2.13 &1.43& 1.13 &--\\\cline{3-9}
         & & Hint in instru. &6.17 &5.3 &2.9 &0.87 &1.60  & --\\\cline{3-9}
         & & no hint & 1.60&2.00 & 8.50& 2.20&9.30 &--\\\cline{3-9}
         & & \enspace\enspace\enspace $E[\mathcal{A}_{w/o}]$  &3  &2.73 &4.51  & 1.5 & 4.01 &  --\\\cline{2-9}
         & \multicolumn{2}{c|}{\newevaluationmetric} &\textbf{36.30} &\textbf{34.24} &\textbf{23.52} &\textbf{20.27} & \textbf{24.56} &--\\\hline
         

       \multirow{6}{*}{EquationInf.}  & \multicolumn{2}{c|}{w/ $\mathcal{G}$ (i.e., $\mathcal{A}_{w/}$)} & 44.71 &55.39 &30.31&20.40&29.90 & --\\\cline{2-9}
         & \multirow{4}{*}{w/o $\mathcal{G}$} & Hint as option & 1.91 &8.67 &38.90&9.06&5.79 &-- \\\cline{3-9}
         & & Hint in instru. &2.86 &9.06 & 0.67&0.19 &0.29  &--\\\cline{3-9}
         & & No hint & 0.0 & 0.0 & 0.0 & 0.0 & 0.0 &--\\\cline{3-9}
         & & \enspace\enspace\enspace $E[\mathcal{A}_{w/o}]$  &1.59  &5.91  &13.19  &3.08  &2.03 &  --\\\cline{2-9}
         & \multicolumn{2}{c|}{\newevaluationmetric} &\textbf{23.15} &\textbf{30.65} &\textbf{21.75} &\textbf{11.74} & \textbf{15.96} &--\\\hline\hline
         
    \end{tabular}
    \caption{\newevaluationmetric~of closed-source, open-source LLMs and humans. We report human performance only on \tasknamemctest~due to its challenging nature and more manageable label size within a reasonable timeframe. Gemma is much lower than other LLMs on \tasknamemctest~ dataset because it cannot follow the instruction and always fail to return the option. }
    \label{tab:EM}
  \vspace{-1em}
\end{table*}

\section{Experiments}\label{sec:experimental_design}

\paragraph{LLM Models.}
We evaluate several popular open-source and closed-source LLMs in this study.

\textbullet\enspace\textbf{Closed-source LLMs.} GPT-4 
 \cite{DBLP:journals/corr/abs-2303-08774} and Claude3 \cite{claude3-anthropic}.

\textbullet\enspace\textbf{Open-source LLMs.} Llama-3 \cite{Llama3-meta}, Gemma \cite{DBLP:journals/corr/abs-2403-08295}, and Mistral \cite{DBLP:journals/corr/abs-2310-06825}.

\paragraph{Experimental setting.}

We run each set of experiments 3 times with options shuffled by different random seeds and report the averaged results. For more details, including model versions, hyper-parameters, and costs, please see Appendix.~\ref{appendix:experiment_details}.

\subsection{Main results}

The main results are presented in Table~\ref{tab:EM}.

\paragraph{$\mathcal{G}$ is present (w/ $\mathcal{G}$).} First, all LLMs, and especially closed-source LLMs, demonstrate exceptionally high accuracy, achieving around 98\% on \tasknamemctest~and over 65\% on \tasknamebanking. Second, the performance of LLMs decreases progressively from \tasknamemctest, \tasknamebanking~to \tasknameequation, which aligns with our expectations due to the increasing level of difficulty of these tasks.

\paragraph{$\mathcal{G}$ is absent (w/o $\mathcal{G}$).} For all three prompt styles in w/o $\mathcal{G}$, LLM performance decreases notably as all models still tend to return one of the incorrect options offered. When comparing hint styles, ``\hintoption'' and ``\hintinstru'', we find that LLMs' preference varies. Such variability indicates the difficulty of consistently evaluating $\mathcal{A}_{w/o}$, as it is somewhat dependent on prompt design. We encourage researchers to design multiple prompts and use the expected value $E[\mathcal{A}_{w/o}]$ rather than any individual $\mathcal{A}_{w/o}^i$.

Llama3 exhibits unexpected behavior on the \tasknameequation~dataset, as ``\hintoption'' performs even better than ``w/ $\mathcal{G}$''.  
We suspect this may be due to data bias during model pretraining. Please see Section \ref{sec:analysis} $\mathcal{Q}_5$ for more analysis.



\paragraph{Observations about \newevaluationmetric.} 
In the last column of Table~\ref{tab:EM}, we report the human performance on \tasknamemctest~(more details in Section \ref{sec:analysis} $\mathcal{Q}_2$). Although both GPT-4 and Claude3 perform on par with humans (98\%+ vs. 100\%), \newevaluationmetric~clearly shows they are still behind humans--both overall (around 80\% by LLMs vs. 97.67\% by humans) and in ``$E[\mathcal{A}_{w/o}]$'' category. 

We can see that when using the standard evaluation approach (with the gold label as an option), some LLMs appear to reach human-like performance. However, our evaluation reveals that LLMs still lag considerably behind human performance because they cannot recognize the absence of the true answer as effectively as humans can. Therefore, \newevaluationmetric~offers a more comprehensive measure to evaluate LLMs' understanding of the classification task and their ability to perform human-level discrimination.

\subsection{Analysis}\label{sec:analysis}

\paragraph{$\mathcal{Q}_1$: Most effective prompt when gold label is missing: \hintoption~, \hintinstru~ or \hintno?}

\hintno~ is clearly the worst among the three. Based on Table \ref{tab:EM}, for strong closed-source LLMs, \hintinstru~consistently results in higher performance than \hintoption. This can be attributed to their superior ability to follow instructions. Additionally, considering the poorer performance of \hintno, it is suggested that when working with widely recognized top-performing LLMs and needing to try only one type of hint (perhaps due to budget constraints), \hintinstru~ is the better option.

Among open-source LLMs, \hintoption~more frequently outperforms \hintinstru, similar to our observations about Llama 3 on \tasknameequation. We suspect this is because open-source LLMs are generally less adept at following instructions, and they might be relying on some specific classification patterns seen during pretraining. Therefore, including \noneofthem~in the instruction is less clear for them compared to setting it as a separate option.



In addition, we notice that these open-source LLMs achieve the highest performance under \hintno~ among the three w/o $\mathcal{G}$ prompts on both \tasknamemctest~ and \tasknamebanking. In fact, open-source LLMs tend to generate a new option whenever the gold option is absent, regardless of whether there are hints about \noneofthem. Therefore, even with \hintoption~ and \hintinstru~, these models often ignore hints and propose self-generated answers without returning \noneofthem. This leads to poorer performance \hintoption~ and \hintinstru.

We also notice interesting differences in the accuracy ranking of LLMs between when the gold label is available (i.e., $\mathcal{A}_{w/}$) and when it is deleted (i.e., $E[\mathcal{A}_{w/o}]$). More details and analysis will be illustrated in Appendix~\ref{appendix:ranking_difference}.


\paragraph{$\mathcal{Q}_2$: Human performance analysis when the gold label is absent}

\begin{figure}[t]
	\begin{center}
		\centering
		\includegraphics[width=0.47\textwidth]{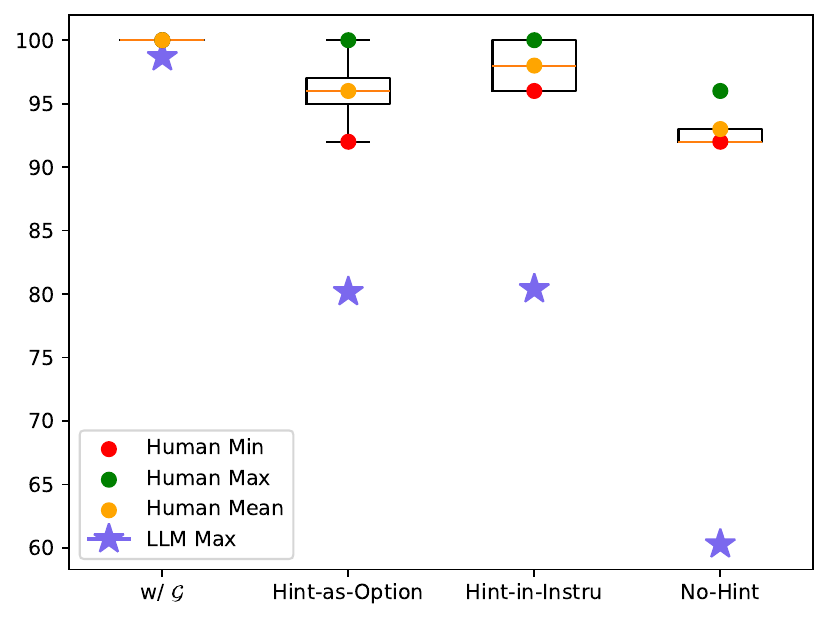}
        \end{center}
  \vspace{-1em}
	\caption{Humans vs. LLMs on \tasknamemctest.}
\vspace{-1em}
\label{fig:human_study}
\end{figure}


In the last column of Table~\ref{tab:EM}, we report the average human performance on \tasknamemctest. 
We randomly selected 25 stories, resulting in 100 questions, each with four answer candidates. We randomly divided the 100 questions into four groups of 25 questions each, with each group corresponding to one of the four prompts (w/$\mathcal{G}$, \hintoption, \hintinstru, or \hintno). We invited four human participants to work separately, with each person responsible for all four groups. To ensure unbiased results, we did not allow the same human to annotate the same question with different prompts, so their annotations of ``w/o $\mathcal{G}$'' questions would not be influenced by ``w/ $\mathcal{G}$'' questions.

Figure \ref{fig:human_study} depicts the statistics of human performance versus the maximum performance of LLMs on 4 prompts. We observe two key dimensions. First, across prompts, human performance is barely affected by the presence of the gold option. In \hintinstru~ and \hintoption, when the gold option is deleted and the ``\noneofthem'' option is provided either in the options or in the instruction, human performance differed by less than 4\% compared to when the gold option was present. Even in \hintno, without any gold option or ``\noneofthem'' option hints, humans were only slightly confused, with the difference being up to 8\%. This indicates that \newetask~ is not a very challenging task for humans, whereas it is for the models. Second, across humans, there is a 4-8\% difference between Human Min and Human Max when the gold option is absent. Despite this, it is evident that even the Human Min performance is significantly higher than the LLM Max performance. Particularly under \hintno, human performance did not show a significant decline compared to other ``w/ $\mathcal{G}$'' prompts, while all LLMs experienced dramatic drops.

\paragraph{$\mathcal{Q}_3$: What different behaviors do LLMs exhibit when the gold option is absent in \hintno?}

\begin{figure}[t]
	\begin{center}
		\centering
		\includegraphics[width=0.47\textwidth]{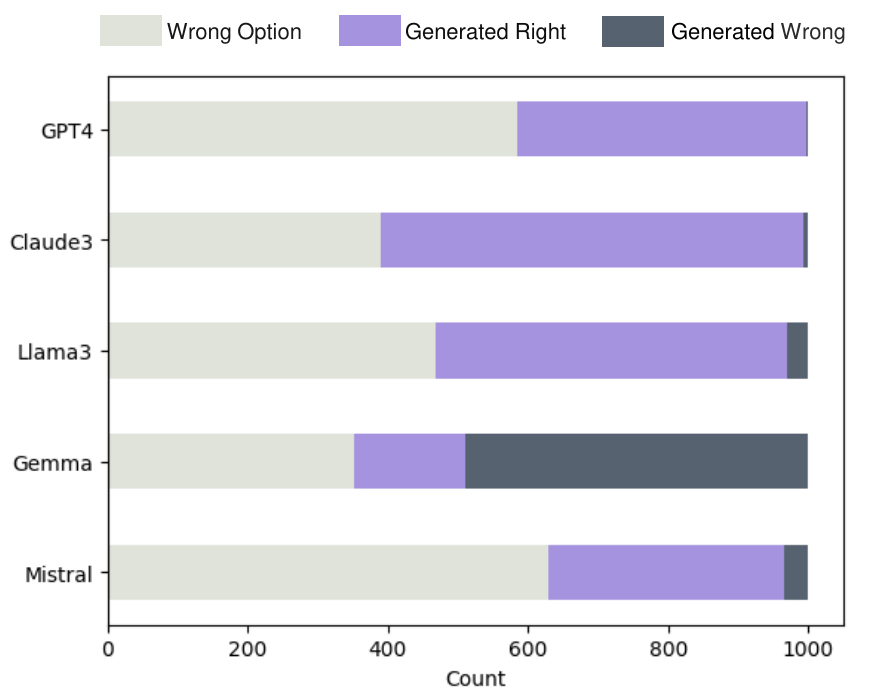}
        \end{center}
\vspace{-1em}
\caption{LLMs' output pattern distribution in \hintno~on \tasknamemctest.}
\label{fig:accu_2_bar_plot}
\vspace{-1.2em}
\end{figure}

In \hintno, there are two types of patterns from models' responses. The first type is to declare that none of the provided options is correct. The second type is to generate a new answer that the model believes to be correct. Each model behaves differently, and their response patterns vary. We identified three behaviors:

\begin{itemize}[leftmargin=10pt]

    \item GPT-4 usually combines both patterns, while other models tend to generate a new answer directly. 

    \item Figure~\ref{fig:accu_2_bar_plot} illustrates the distribution of LLM responses in \hintno~ on \tasknamemctest. GPT-4 and Mistral make the most mistakes on incorrect options, while Claude3 and Llama 3 are most likely to generate correct new labels. Gemma performs the worst, with mostly wrong labels generated.

    \item When generating new answers, a significant difference lies in the letter assigned by LLMs to this new option. We observed that when only options A-C exist in \hintno, GPT-4 consistently labels its self-generated answer as ``D'' while other LLMs tend to choose letters from A to C. 
\end{itemize}

Sample outputs and further analysis of behavior patterns are available in Appendix~\ref{appendix:no_hint_model}.


\paragraph{$\mathcal{Q}_4$: If LLMs can generate a correct label in \hintno, is it because the LLMs have seen this dataset during pretraining? }

When eyeballing the answers suggested by LLMs in \hintno~ for \tasknamemctest~ and \tasknamebanking, we noticed that the model sometimes generates a new option identical to the gold option. This is not surprising for \tasknamemctest, where options can be direct terms from the story. However, for \tasknamebanking, with its fixed and specific label space, this raises questions about whether LLMs have had this dataset in its parameterized knowledge base.

To answer this question, we employed a small trick with the \tasknamebanking~ dataset. Given that the tokens in \tasknamebanking's options are delimited by `` \_ '', such as ``lost\_or\_stolen\_card'', we replaced `` \_'' with ``-'' and reran the experiments for the \hintno~ scenario. Among our five LLMs, Llama-3 was the only model that still generated ``\_'' in the output. Therefore, we highly suspect that Llama-3 was exposed to \tasknamebanking~during its pretraining. Consequently, its performance may be biased, especially in the human evaluation metrics.
    
Even though all other LLMs consistently follow the new delimiter ``-'', this can be attributed to their strong instruction-following capabilities. Therefore, our proposed format-aware trick is unable to conclusively determine whether these models were exposed to this dataset during pre-training.

\paragraph{$\mathcal{Q}_5$: Would the model be misled when we add \noneofthem~ in w/ $\mathcal{G}$?} We have observed that the model exhibits different behaviors when encountering \noneofthem~ hints. We hypothesize that this behavior might stem from data bias introduced during model pretraining or instruction tuning. To investigate this, we conducted an ablation study by introducing \noneofthem~ options in w/ $\mathcal{G}$ prompts on a subset (250 instances) on \tasknamemctest~ and \tasknameequation.

To ensure fairness, we randomly replaced one incorrect answer with \noneofthem, ensuring the models always select from options A-D in both scenarios. The results, presented in Table~\ref{tab:with_gold_none_listed}, show that for the simpler \tasknamemctest~ dataset, model performance remains nearly identical between w/ $\mathcal{G}$ and w/ $\mathcal{G} + None$ settings. For the more challenging \tasknameequation~ dataset, closed-source models maintains their robustness, while open-source models, particularly Llama 3 and Gemma, experienced significant performance declines. This decrease is attributed to the confusion caused by the \noneofthem~ option, which might also explain why Llama 3 performs exceptionally well in \hintoption~ on \tasknameequation.

\begin{table}[t]
\centering
\small
\setlength{\tabcolsep}{2pt}
    \begin{tabular}{c c|c|c|c|c|c}
        \hline
        & & GPT 4 & Claude 3 & Llama 3 & Gemma & Mistral \\
        \hline
        \hline
        \multirow{2}{*}{\textsc{MC}} & w/ $\mathcal{G}$ & 98.67 & 98.26 & 94.23 & 39.00 & 87.53 \\
        & + None & 99.60 & 97.60 & 91.70 & 40.00 & 87.60 \\
        \hline
        \multirow{2}{*}{\textsc{Eq}} & w/ $\mathcal{G}$ & 44.71 & 55.39 & 30.31 & 20.40 & 29.90 \\
        & + None & 45.60 & 58.80 & 15.30 & 10.10 & 22.40  \\
        \hline
    \end{tabular}
\vspace{-0.5em}
    \caption{Ablation study: w/ $\mathcal{G}$ vs w/ $\mathcal{G} + None$. \textsc{MC} and \textsc{Eq} refer to \tasknamemctest~ and \tasknameequation, respectively.}
    \label{tab:with_gold_none_listed}
\vspace{-1em}
\end{table}

\section{Conclusion}
Our study reveals critical insights into the limitations of LLMs in classification tasks under \newetask~ where gold labels can be present or absent. The \taskname~ benchmark and  \newevaluationmetric~ metrics provide a comprehensive evaluation by combining metrics for both the presence and absence of gold labels. This work establishes a new testbed for assessing LLMs' human-level discrimination intelligence, offering a framework for future research aimed at improving the robustness and reliability of LLMs. 

\section*{Limitation}
There are several limitations in \taskname~ as presented. First, we tested only three different prompts in w/o 
$\mathcal{G}$. There are many other possible prompts that could assess the model's ability without the gold label, but it is impossible to address them all in this paper. Second, the \tasknameequation~ dataset requires a large number of input tokens due to its problem setting, making it expensive for proprietary models. However, researchers are welcome to test \taskname~ on any open-source models or adapt \newevaluationmetric~ to any dataset, not limited to the models or datasets used in this work.

\bibliography{anthology, custom}

\appendix
\section{Appendix}

\subsection{Scaling \tasknameequation}
\label{appendix:equation_scale}

\begin{figure}[t]
	\begin{center}
		\centering
		\includegraphics[width=0.45\textwidth]{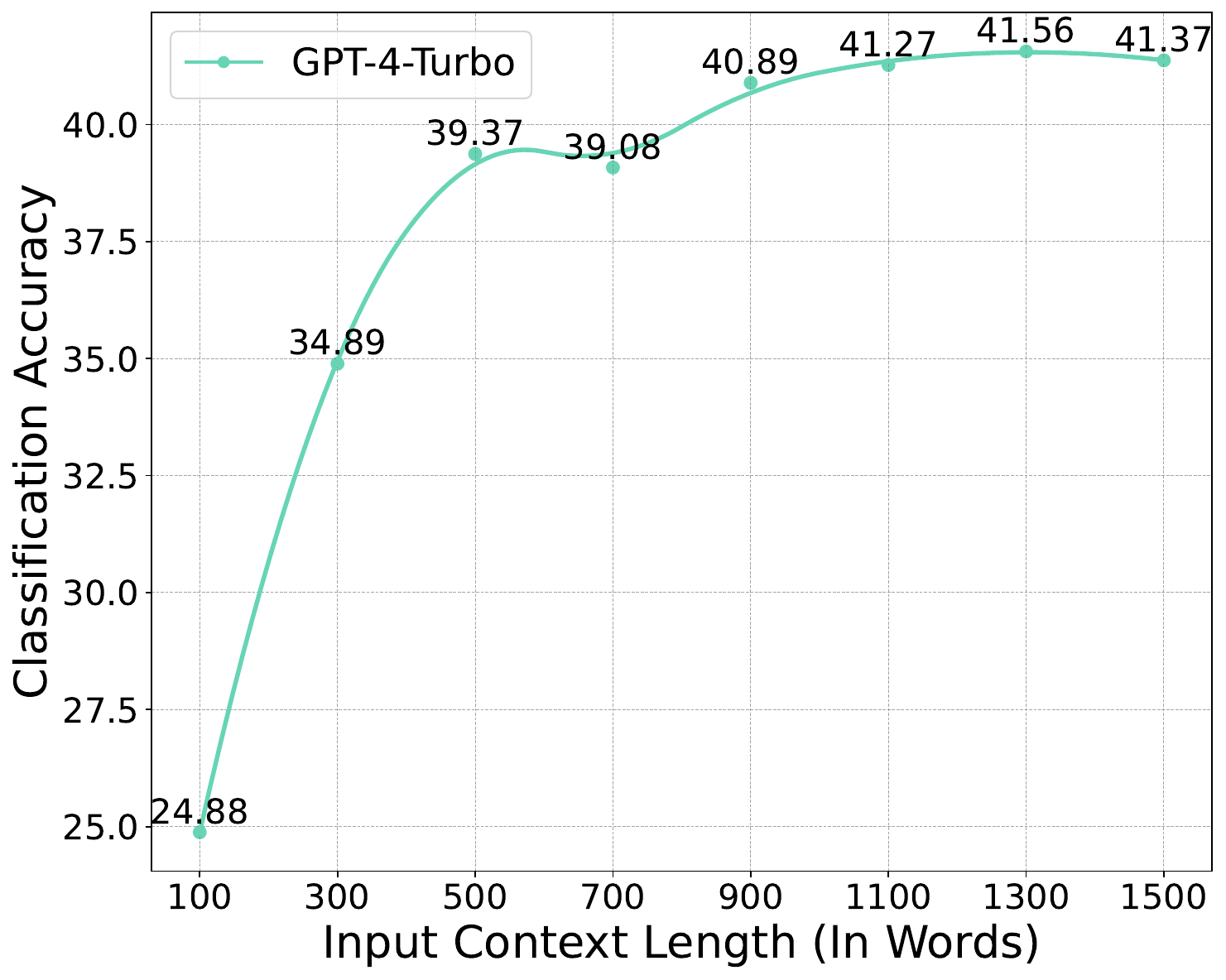}
        \end{center}
	\caption{Scaling the length of context around the equation in \tasknameequation}
\label{fig:equation_input_context_scaling}
\vspace{0em}
\end{figure}

To keep an optimal context length for both sides of the equation, we tested ten different context lengths ranging from 100 to 1500 on GPT-4. The results are displayed in Figure \ref{fig:equation_input_context_scaling}. We found that starting from 1000 words, the model's performance did not show significant improvement. Therefore, we decided to retain 1000 words for either side. Additionally, when truncating the context, we ensure that complete sentences are presented. This means if the 1000-word limit would cut off a sentence, we extend the truncation to ensure sentence completeness.

\subsection{Prompt Illustrations}
\label{appendix:prompt_example}
Below, we present sample inputs from the \tasknamemctest, \tasknamebanking, and \tasknameequation~ datasets. These examples represent w/ $\mathcal{G}$ prompts, where the gold option is present (in blue). For w/o $\mathcal{G}$ prompts, the gold option is removed, and the prompt is adjusted accordingly. Specifically, we add hints (\noneofthem) in options/instruction for \hintoption/\hintinstru, or no hint at all for \hintno.

\textbf{\tasknamemctest} \\
\noindent
\fbox{%
 \begin{minipage}{1\linewidth} 

\textbf{Task:} \\
Given the story and an associated question, please return the correct option for the question without explanation. \\
\\
\textbf{Story:} \\
\textbf{[$\cdots\cdots\cdots$]}I'm going to play for the Yankees ma!" Tom said. Tom's mom was so excited that she took Tom and the whole family out for dinner. Grandpa, Grandma, Mom and Dad were all there, and bought Tom a big cake! \textbf{[$\cdots\cdots\cdots$]}\\
\\
\textbf{Question:} \\
What did Tom's family buy him to celebrate?\\
\\
\textbf{Options:} \\
    {\color{blue}A: A cake}\\
    B: A car\\
    C: New clothes \\
    D: A baseball  \\
\\
\textbf{Your answer:} \\
    \end{minipage}
}

\textbf{\tasknamebanking} \\
\noindent
\fbox{%
 \begin{minipage}{1\linewidth} 

\textbf{Task:} \\
You are going to perform an intent classification task. Given an utterance and multiple intent class options, return the correct intent class name only.\\
\\
\textbf{Utterance:} \\
An unauthorized payment is in my app\\
\\
\textbf{Intent Class Options:} \\
refund\_not\_showing\_up, apple\_pay\_or\_google\_pay, \\
pending\_card\_payment, {\color{blue}card\_payment\_not\_recognised}, \\
\textbf{[$\cdots\cdots\cdots$]} \\
balance\_not\_updated\_after\_cheque\_or\_cash\_deposit\\
\\
\textbf{Your answer:} \\
    \end{minipage}
}

\textbf{\tasknameequation} \\
\noindent
\fbox{%
 \begin{minipage}{1\linewidth}

\textbf{Task:} \\
You are given the latex source code of the context before and after an equation in an NLP paper and multiple options for the equation. Only return the correct option letter as the answer without explanation.\\
\\
\textbf{Context before:} \\
``[$\cdots\cdots\cdots$]$\bm{e}_{x^i}$ and $\bm{e}_{z^i}$ are the average representations along the sequence dimension from the encoder outputs. Apart from the contrastive loss, the standard cross-entropy loss is calculated as:''
\\
\\
\textbf{Context after:} \\
``We combine both losses as the final loss:
\begin{align}
\mathcal{L} = \mathcal{L}_{ce} + \lambda \mathcal{L}_{ctr} \nonumber
\end{align}
where $\lambda$ is an interpolation factor. [$\cdots\cdots\cdots$]''
\\
\\
\textbf{Equation options:} \\
    A: $\mathcal{L}_{ce} = -\sum_{i=1}^{N} \sum_{c=1}^{C} y_{i,c} \log(p_{i,c})$ \\
    {\color{blue}B: $\mathcal{L}_{ce} = - \sum _{i=1}^{N} (\log P_{\theta}(y^i | x^i) + \log P_{\theta}(y^i | z^i))$} \\
    C: $\mathcal{L}_{ce}  = - \frac{1}{N} \sum_{i=1}^{N} \sum_{c=1}^{C} y^i_c \log(p^i_c)$ \\
    D: $\mathcal{L}_{ce} = -\sum_{i=1}^N \sum_{k=1}^C y_k^i \log \hat{y}_k^i$ 
\\
\\
\textbf{Your answer:} \\
\end{minipage}
}

\subsection{Experimental Setting Details}
\label{appendix:experiment_details}

We adopt the most recently released instruction-tuned versions of each LLM model as follows: \textit{gpt-4o-2024-05-13}, \textit{claude-3-opus-20240229}, \textit{Meta-Llama-3-8B-Instruct}, \textit{gemma-7b-it}, \textit{Mistral-7B-Instruct-v0.2}. During inference, the temperature is always set to 0 for reproducibility. The top\_p is set to 0.9, and the max\_length for generation is 1000. 

The cost of using the two closed-source LLMs is detailed below. For \tasknamebanking, \tasknamemctest, and \tasknameequation, it takes 1 million, 0.4 million, and 5 million input tokens, respectively, to run one prompt on all instances. Given that one set of experiments includes 4 prompts, the costs for GPT-4 (including both input and output tokens) are approximately \$15, \$8, and \$100 for \tasknamebanking, \tasknamemctest, and \tasknameequation, respectively. For Claude 3, the costs are roughly \$60, \$24, and \$300 for \tasknamebanking, \tasknamemctest, and \tasknameequation, respectively.

We want to highlight that \taskname~ and \newevaluationmetric~ represent a novel evaluation approach under \newetask. They can be applied to any model in any classification setting, extending beyond the three datasets and five models reported in this study.

\subsection{Ranking Differences of LLMs between $\mathcal{A}_{w/}$ and $\mathcal{A}_{w/o}$}
\label{appendix:ranking_difference}

Are there any differences in the accuracy ranking of LLMs between when the gold label is available (i.e., $\mathcal{A}_{w/}$) and when it is deleted (i.e., $E[\mathcal{A}_{w/o}]$)? We notice that closed-source models consistently achieve impressive accuracy when the gold option is available. However, when the gold option is removed, they resist acknowledging the absence of a correct label, especially in more challenging tasks. In \tasknamemctest, the simplest of the three tasks, GPT-4 and Claude 3 confidently suggest ``none'' or generate a new answer, outperforming other models by a large margin. Conversely, for \tasknamebanking~ and \tasknameequation, they tend to select from the available incorrect options, performing as poorly as the open-source models.

\subsection{Model Behaviour under \hintno}
\label{appendix:no_hint_model}
In this section, we provide a detailed analysis of each model's responses in \hintno. Since \tasknamemctest~ is the simplest of the three tasks, it is the most likely scenario for the model to correctly identify the \hintno~ context. In contrast, for \tasknameequation~ and \tasknamebanking, it is very rare for the model to identify the situation. Hence, we will focus on \tasknamemctest~ for our primary discussion.

Even with \tasknamemctest, models fail to recognize the absence of the gold label approximately half of the time. When this occurs, the typical response patterns are as follows:\\ 
\fbox{%
 \begin{minipage}{1\linewidth}
\noindent
\textbf{Gold option:} Rick\\
\textbf{Options:} A. Bob, B. James, C. Stephanie\\
\textbf{\textit{Response type 1:}} \textit{``None of the above''} \\
\textbf{\textit{Response type 2:}} \textit{``None of the options provided are correct. The correct answer is Rick.''} \\
\textbf{\textit{Response type 3:}}  ``\textit{D. Rick''}\\
\textbf{\textit{Response type 4:}}  ``\textit{C. Rick''}\\
\end{minipage}
}

Each model displays these patterns differently, with unique characteristics specific to each model.

\paragraph{GPT-4}
Surprisingly, GPT-4 is among the least likely models to acknowledge the absence of a gold label. However, when it does recognize this situation and adopts an ``out-of-the-box'' approach, its accuracy is notably the highest. The most common responses for GPT-4 are \textbf{\textit{Response types 1, 2, and 3}}. For the new answers proposed, GPT-4 always labels them as option D.

\paragraph{Claude 3 / Llama 3 / Mistral}
Claude 3, Llama 3, and Mistral exhibit consistent and standardized response patterns in the format ``\textit{\textbf{Letter}. Option}'', without giving any additional discussion. The option might be one of the incorrect options provided or a new answer generated by the model. Interestingly, unlike GPT-4, which always assigns its new answer as option D, Claude 3, Mistral, and Llama 3 tend to assign a letter from A to C, as per \textbf{\textit{Response type 4}}.

\paragraph{Gemma}
Gemma's responses are the most chaotic and illogical. It often forces an explanation on the incorrect option or provides an incorrect option but includes the correct answer in the reasoning. This inconsistency aligns with its poor performance across all datasets and most prompts. One example of such an error is as follows:\\
\noindent
\textbf{Question:}What color was the animal's stripe? \\
\textbf{Gold option:} green\\
\textbf{Options:} A. playground, B. sand, C. frosting \\
\textbf{Gemma output:}\\
\textit{``Answer: C\\
The animal had a green stripe across its back, therefore the answer is C.''}\\
As a result, it is very rare for Gemma to generate correct output, and none of the \textbf{\textit{Response types}} fit its behavior.

\subsection{Human Behaviour under \hintno}
\label{appendix:no_hint_human}

\fbox{%
 \begin{minipage}{1\linewidth}
\noindent
\textbf{Gold option:} Rick\\
\textbf{Options:} A. Bob, B. James, C. Stephanie\\
\textbf{\textit{Response type 1:}} \textit{``None of the options is correct /  something is wrong with the problem setting''} \\
\textbf{\textit{Response type 2:}} \textit{``Rick''} \\
\end{minipage}
}

When faced with \hintno, humans respond more straightforwardly and directly compared to the models. Humans often either point out there might be flaws in the question design or provide the correct answer directly as the two \textbf{\textit{response types}} above. Very interestingly, we notice that humans would not assign a letter to a self-generated answer and treat it as one of the provided options. This behavior seems to be unique to models, likely because they are trained to follow the provided pattern.

\subsection{ACL ethics code discussion}
\label{appendix:ACL_ethics}

\textbullet\enspace\textbf{Time/Memory Cost} We only collect inferences from LLMs. For closed-source models (GPT-4 and Claude 3) accessed through API, there's no significant memory usage. For open-source models (Llama 3, Gemma and Mistral) accessed through Huggingface, the memory usage is the same as the model parameter size. It takes about an hour on average to run all experiments on one dataset.

\textbullet\enspace\textbf{Scientific artifacts usage} The existing Scientific artifacts included in this work are 5 models (GPT-4, Claude 3, Llama 3, Gemma and Mistral. please refer to Section \ref{sec:experimental_design}. LLM Models) and 3 NLP classification datasets (\tasknamemctest, \tasknamebanking~ and \tasknameequation, please refer to Section \ref{sec:datasets}). The models and datasets used in this work are publicly available for research purposes and do not contain any sensitive information. Our use of existing Scientific artifacts is consistent with their intended usage. The dataset \tasknameequation~ proposed by us does not contain any personal information. 

The license, copyright information, the asset we proposed, and terms of use information regarding \taskname, will be specified once the code is released.

\end{document}